\documentclass[preprint,12pt,3p]{elsarticle}

\usepackage{amssymb}
\usepackage[cmex10]{amsmath}
\usepackage{bm}
\usepackage{bbm}
\usepackage{graphicx}
\graphicspath{{./img}}
\usepackage{subcaption}
\usepackage{textcomp}
\usepackage{url}
\usepackage{lipsum}

\makeatletter
\def\ps@pprintTitle{%
   \let\@oddhead\@empty
   \let\@evenhead\@empty
   \let\@oddfoot\@empty
   \let\@evenfoot\@oddfoot
}
\makeatother

\begin{document}

\begin{frontmatter}

\title{Neural networks for topology optimization}

\author[uva,sk]{Ivan Sosnovik}
\ead{i.sosnovik@uva.nl}
\author[sk,inm]{Ivan Oseledets}
\ead{i.oseledets@skoltech.ru}

\address[uva]{University of Amsterdam, Amsterdam, The Netherlands}
\address[sk]{Skolkovo Institute of Science and Technology, Moscow, Russia}
\address[inm]{Institute of Numerical Mathematics RAS, Moscow, Russia.}

\begin{abstract}
In this research, we propose a deep learning based approach for speeding up the topology optimization methods. The problem we seek to solve is the layout problem. The main novelty of this work is to state the problem as an image segmentation task. We leverage the power of deep learning methods as the efficient pixel-wise image labeling technique to perform the topology optimization. We introduce convolutional encoder-decoder architecture and the overall approach of solving the above-described problem with high performance. The conducted experiments demonstrate the significant acceleration of the optimization process. The proposed approach has excellent generalization properties. We demonstrate the ability of the application of the proposed model to other problems. The successful results, as well as the drawbacks of the current method, are discussed.
\end{abstract}

\end{frontmatter}

\section{Introduction}
\label{sec:intro}
Topology optimization solves the layout problem with the following formulation: how to distribute the material inside a design domain such that the obtained structure has optimal properties and satisfies the prescribed constraints? The most challenging formulation of the problem requires the solution to be binary i.e. it should state whether there is a material or a void for each of the parts of the design domain. One of the common examples of such an optimization is the minimization of elastic strain energy of a body for a given total weight and boundary conditions. Initiated by the demands of automotive and aerospace industry in the $20^{th}$ century, topology optimization has spread its application to a wide range of other disciplines: e.g. fluids, acoustics, electromagnetics, optics and combinations thereof \cite{topopt_application}.

All modern approaches for topology optimization used in commercial and academic software are based on finite element methods. SIMP (Simplified Isotropic Material with Penalization), which was introduced in 1989 \cite{Bendsoe1989}, is currently a widely spread simple and efficient technique. It proposes to use penalization of the intermediate values of density of the material, which improves the convergence of the solution to binary. Topology optimization problem could be solved by using BESO (Bi-directional evolutionary structural optimization) \cite{beso_1, beso_2} as an alternative. The key idea of this method is to remove the material where the stress is the lowest and add material where the stress is higher. The more detailed review is discussed in Section \ref{sec:topopt}.

For all of the above-described methods, the process of optimization could be roughly divided into two stages: general redistribution of the material and the refinement. During the first one, the material layout varies a lot from iteration to iteration. While during the second stage the material distribution converges to the final result. The global structure remains unchanged and only local alteration could be observed. 

In this paper, we propose a deep learning based approach for speeding up the most time-consuming part of a traditional topology optimization methods. The main novelty of this work is to state the problem as an image segmentation task. We leverage the power of deep learning methods as an efficient pixel-wise image labeling technique to accelerate modern topology optimization solvers. The key features of our approach are the following:

\begin{itemize}
    \item acceleration of optimization process;
    \item excellent generalization properties;
    \item absolutely scalable techniques; 
\end{itemize}


\section{Topology Optimization Problem}
\label{sec:topopt}
Current research is devoted to topology optimization of mechanical structures. Consider a design domain $\Omega : \{\omega_j\}_{j=1}^N$, filled with
a linear isotropic elastic material and discretized with square finite elements. The material distribution is described by the binary density variable $x_j$ that represents either absence (0) or presence (1) of the material at each point of the design domain. Therefore, the problem, that we seek to solve, can be written in mathematical form as:
\begin{equation}\label{nn_topopt:mechanical_problem}
\begin{split}
	\begin{cases}
		\min\limits_{\bm{x}} & 
        c(\bm{u}(\bm{x}), \bm{x}) = 
        \sum\limits_{j=1}^{N}
        E_j(x_j)\bm{u}^T_j \bm{k_0} \bm{u}_j\\
        \hfill \text{s.t.} & 
        V(\bm{x}) / V_0 = f_0 \\
        \hfill & \bm{KU} = \bm{F} \\
        \hfill & x_j \in \{ 0; 1\}, 
        \quad  j  = 1 \dots N
    \end{cases}
    \end{split}
\end{equation}
where $c$ is a compliance, $\bm{u_j}$ is the element displacement vector, $\bm{k_0}$ is the element stiffness matrix for an element with unit Young’s modulus, $\bm{U}$ and $\bm{F}$ are the  global displacement and force vectors, respectively and  $\bm{K}$ is  the global stiffness matrix.  $V(\bm{x})$ and  $V_0$ are the material volume and design domain volume, respectively. $f_0$ is the prescribed volume fraction.

The discrete nature of the problem makes it difficult to solve.  Therefore, the last constraint in (\ref{nn_topopt:mechanical_problem}) is replaced with the following one: $x_j \in [ 0; 1], \;  j  = 1 \dots N$. The most common method for topology optimization problem with continuous design variables is so-called SIMP or power-law approach \cite{Bendsoe1989,Mlejnek1992}. This is a gradient-based iterative method with the penalization of non-binary solutions, which is achieved by choosing Young\textquotesingle s modulus of a simple but very efficient form:
\begin{equation}\label{nn_topopt:simp_young}
E_j(x_j) = E_{\text{min}} +
x^p_j (E_0  - E_{\text{min}})
\end{equation}

The exact implementation of SIMP algorithm is out of the scope of the current paper. The updating schemes, as well as different heuristics,  can be found in  excellent  papers \cite{bendsoe1995optimization,sigmund1997,bourdin2001,Svanberg2013,Groenwold2009}. The topology optimization code in Matlab is described in details in  \cite{topopt99lines,topopt88lines} and the Python implementation of SIMP algorithm is represented in \cite{topy}.

\begin{figure}
	\centering
	\includegraphics[width=0.5\linewidth]
    {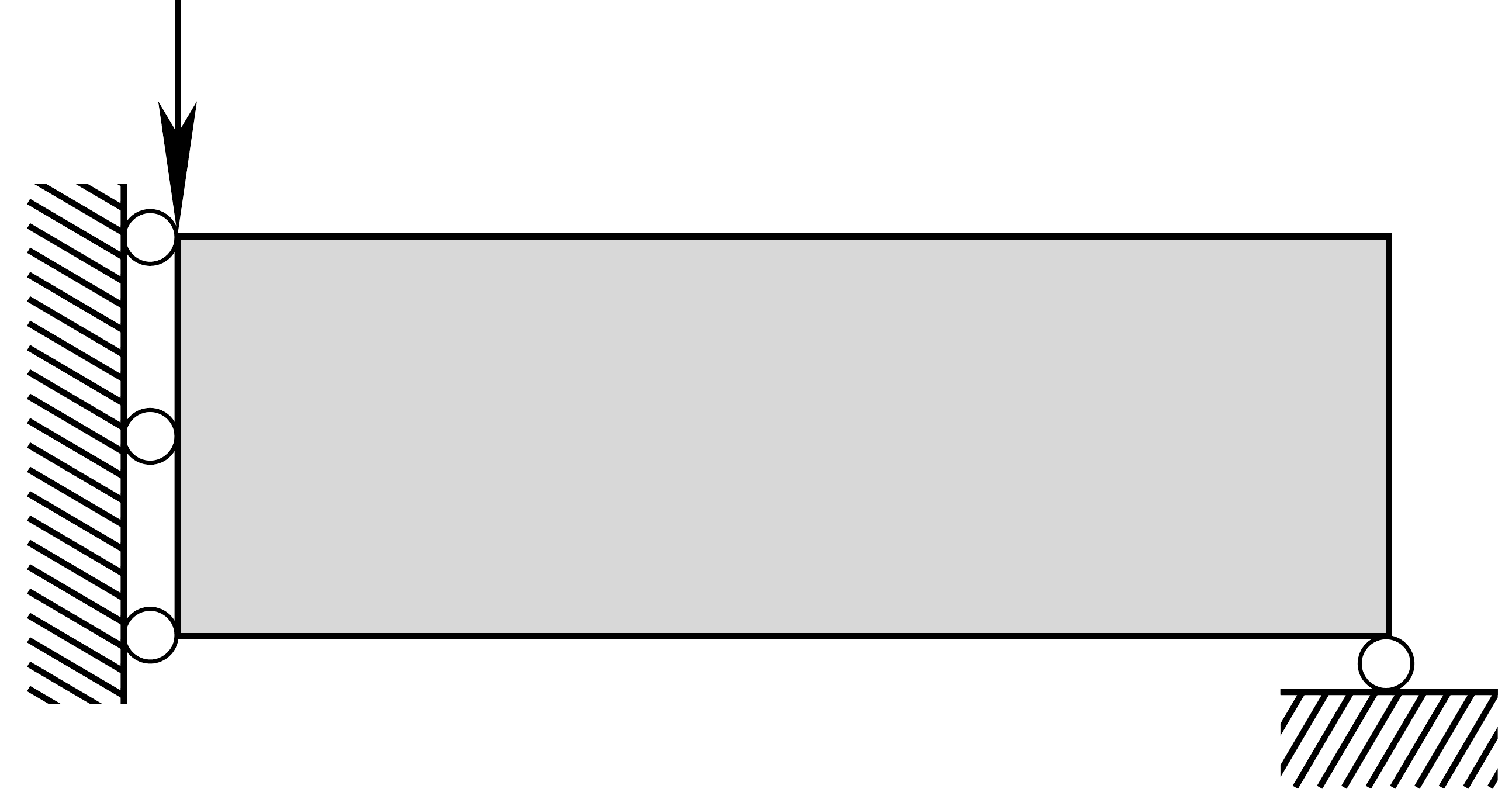}
    \caption{The design domain, boundary conditions, and external load for the optimization of a half MBB beam.}
\label{nn_topopt:mbb_beam}
\end{figure}

\begin{figure}
    \centering
    \begin{subfigure}[b]{0.4\textwidth}
        \includegraphics[width=\textwidth]
        {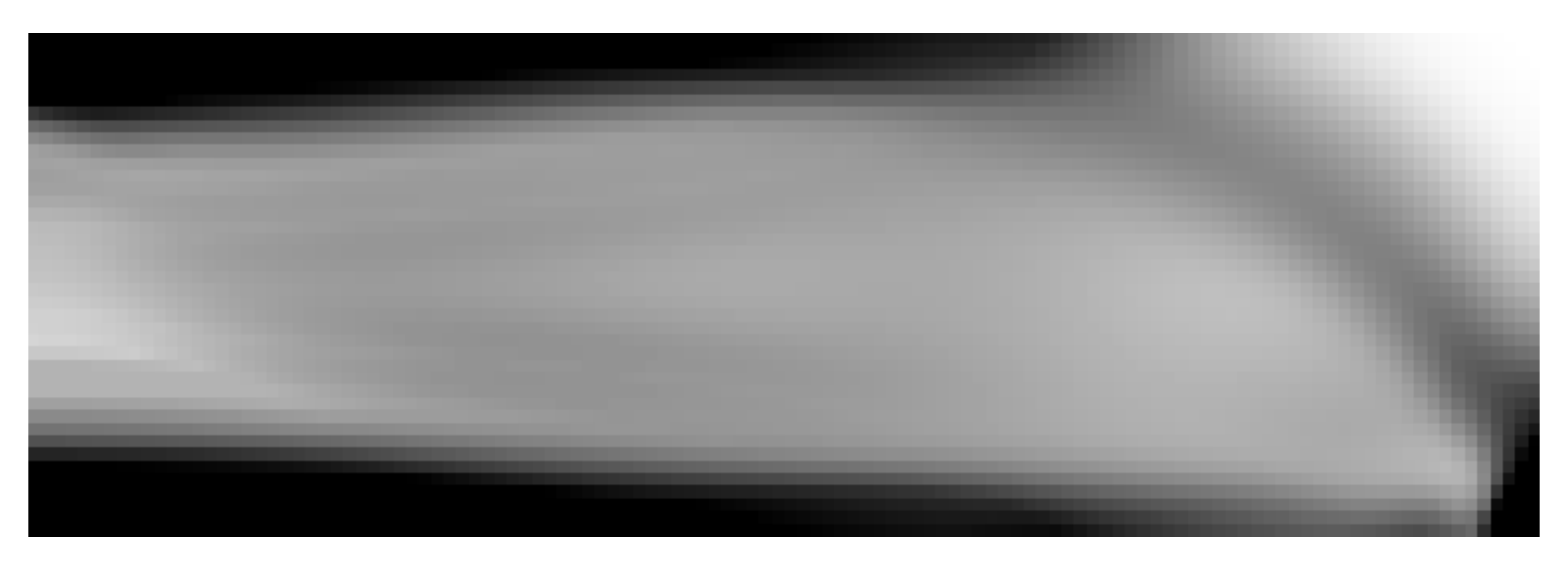}
        \caption{Iteration 3}
    \end{subfigure}
    \begin{subfigure}[b]{0.4\textwidth}
        \includegraphics[width=\textwidth]
        {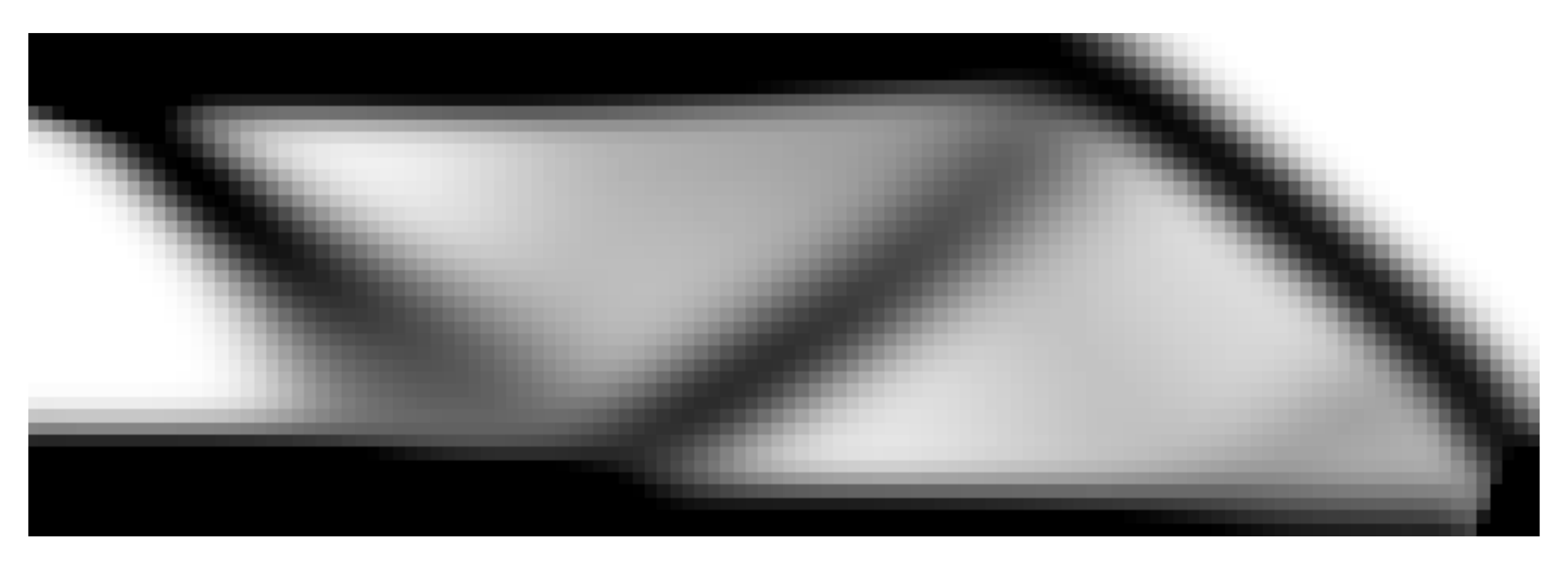}
        \caption{Iteration 13}
    \end{subfigure}
    \begin{subfigure}[b]{0.4\textwidth}
        \includegraphics[width=\textwidth]
        {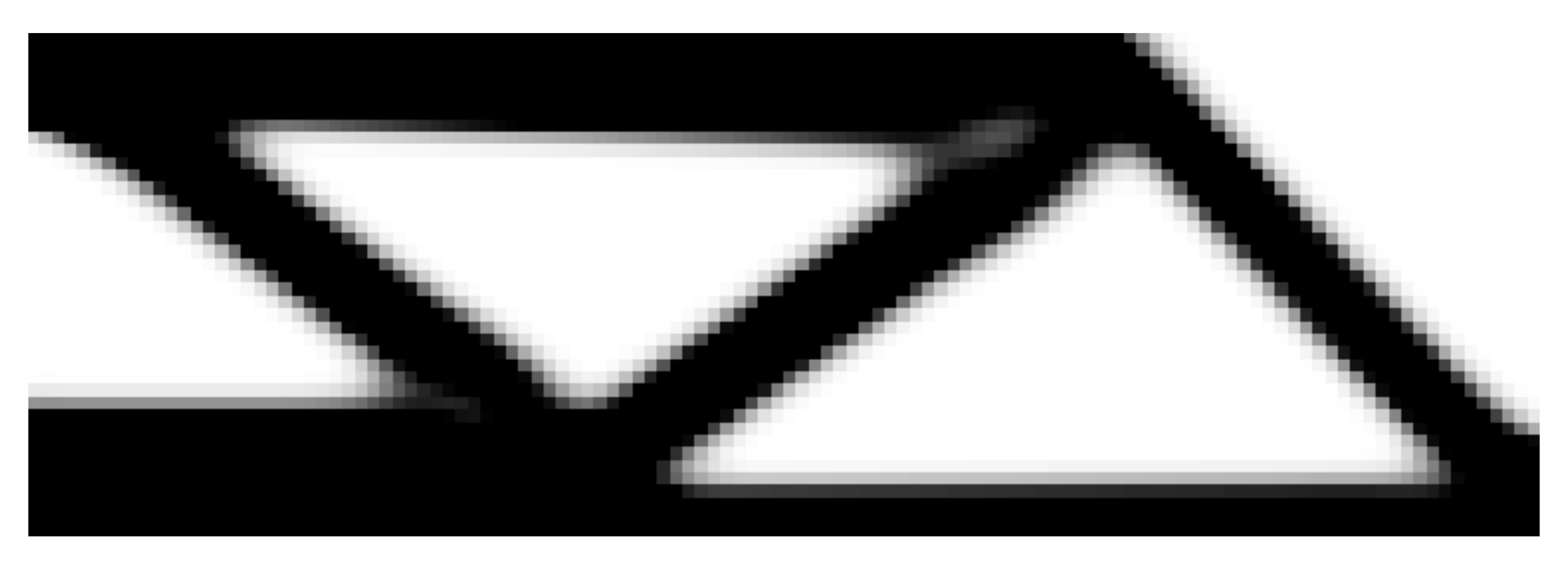}
        \caption{Iteration 30}
    \end{subfigure}
    \begin{subfigure}[b]{0.4\textwidth}
        \includegraphics[width=\textwidth]
        {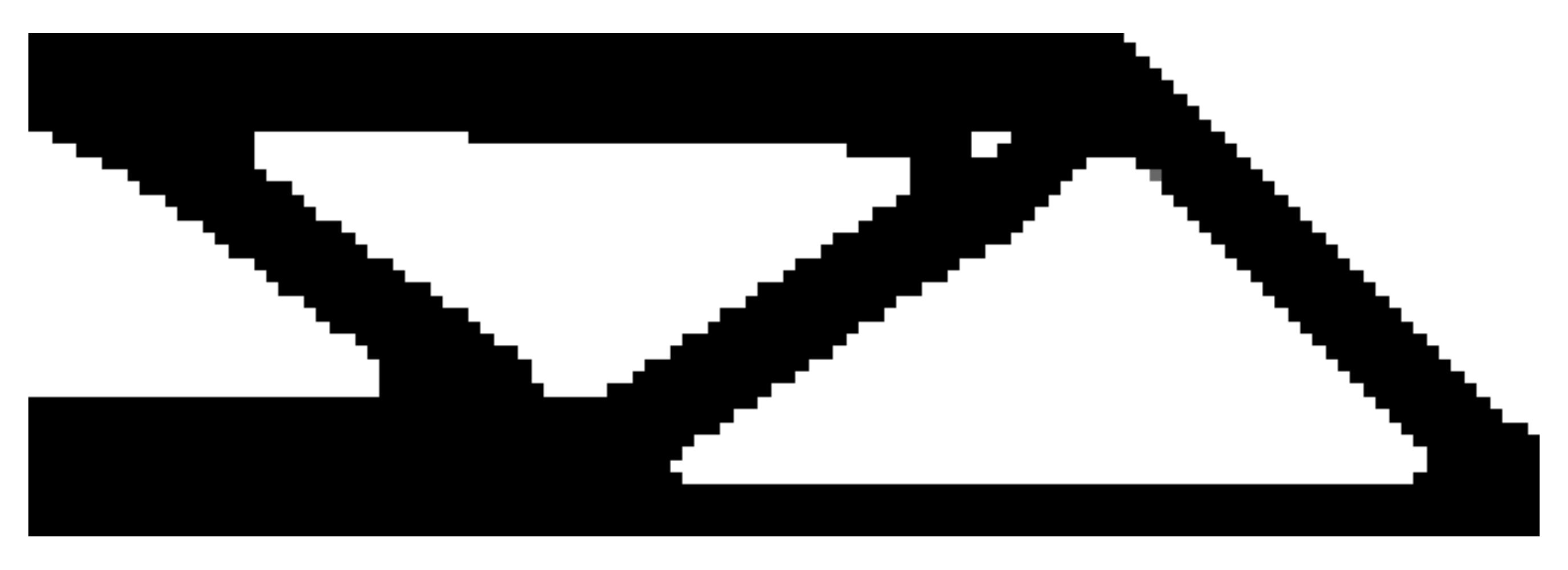}
        \caption{Iteration 80}
    \end{subfigure}
    \caption{The process of optimization of half MBB beam with SIMP method. 120 $\times$ 40 mesh. Black --- 1, white --- 0}
\label{nn_topopt:simp_iters}
\end{figure}

Standard half MBB-Beam problem is used to illustrate the process of topology optimization. The design domain, constraints, and loads are represented in Figure \ref{nn_topopt:mbb_beam}. The optimization of this problem is demonstrated in Figure \ref{nn_topopt:simp_iters}. During the initial iterations, the general redistribution of the material inside of the design domain is performed. The rest of the optimization process includes the filtering of the pixels: the densities with intermediate values converge to binary values and the  silhouette of the obtained structure remains almost unchanged.  


\section{Learning Topology Optimization}
\label{sec:learn_topopt}

As it was illustrated in Section \ref{sec:topopt}, it is enough for the solver to perform a few number $N_0$ of iterations to obtain the preliminary  view of a structure.  The fraction of non-binary  densities could be close to 1, however, the global layout pattern is close to the final one. The obtained image $I$ could be interpreted as a blurred image of a final structure, or an image distorted by other factors. The  thing is that there are just two types of objects on this image: the material and the void. The image $I^*$, obtained as a result of topology optimization does not contain intermediate values and, therefore, could be interpreted as the mask of image $I$. According to this notation, starting from iteration $N_0$ the process of optimization $I \rightarrow I^*$ mimics the process of image segmentation for two classes or foreground-background segmentation. 

We propose the following pipeline for topology optimization: use SIMP method to perform the initial iterations and get the  distribution with non-binary densities; use the neural network to perform the segmentation of the obtained image and converge the distribution to $\{0, 1\}$ solution.

\subsection{Architecture}
\begin{figure}
    \centering
    \includegraphics[width=\textwidth]
    {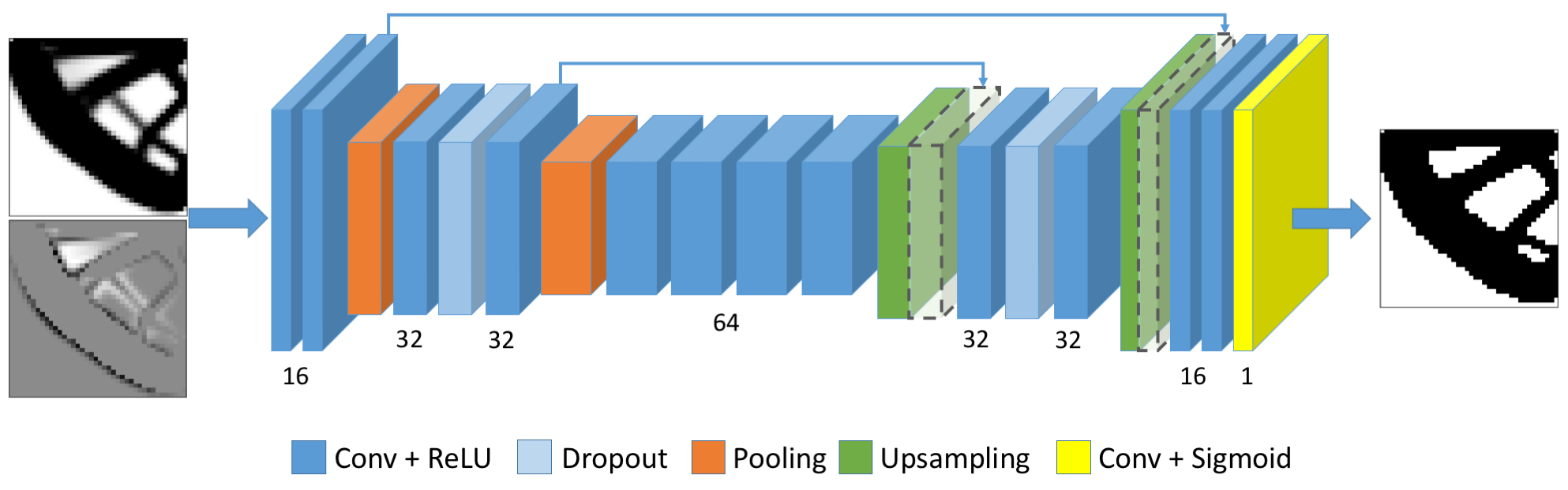}
    \caption[Architecture of the neural network for topology optimization]{The architecture of the proposed neural network for topology optimization. All kernels are of size $3 \times 3$. The number of kernels is represented by the number at the bottom of the layer. Blue arrows and opaque boxes represent the concatenation of the features from different layers.}
\label{fig:nn_arch}
\end{figure}

Here we introduce the \textbf{\textit{Neural Network for Topology Optimization}} --- deep fully-convolutional neural network aimed to perform the  convergence of densities during the topology optimization process. The input of the model is two grayscale images (or a two-channel image). The first one is the density distribution $X_n$ inside of the design domain which was obtained after the last performed iteration of topology optimization solver. The second input is the last performed update (gradient) of the densities $\delta X = X_n - X_{n-1}$, i.e. the difference between the densities after the $n$-th iteration and $n-1$-th iteration.  The output of the proposed model is a grayscale image of the same resolution as an input, which represents the predicted final structure. The architecture of our model mimics the common for the image segmentation hourglass shape. The proposed model has an encoder network and a corresponding decoder network, followed by a final pixel-wise classification layer. The architecture is illustrated in Figure \ref{fig:nn_arch}. 

The encoder network consists of 6 convolutional layers. Each layer has kernels of size $3 \times 3$ and is followed by ReLU nonlinearity. The first two layers have 16 convolutional kernels. This block is followed by the pooling of the maximal element from the window of size $2 \times 2$. The next two layers have 32 kernels and are also followed by MaxPooling layer. The last block consists of 2 layers with 64 kernels each. 

The decoder network copies the architecture of the encoder part and reverses it. The MaxPooling layers are replaced with Upsampling layers followed by the concatenation with features from the corresponding low-level layer as it is performed in U-Net \cite{RonnebergerU-Net:Segmentation}. The pooling operation introduces the invariance of the subsequent network to small translations of the input. The concatenation of features from different layers allows one to benefit from the use of both the raw low-level representation and significantly encoded parametrization from the higher levels. The decoder is followed by the Convolutional layer with 1 kernel and sigmoid activation function. We included 2 Dropout layers \cite{dropout} as the regularization for our network. 

The width and height of the input image could vary, however, they must be divisible by 4 in order to guarantee the coherence of the shapes of tensors in the computational graph. The proposed neural network has just 192,113 parameters. 

\subsection{Dataset}
\label{sect:dataset}

To train the above-described model, we need example solutions to the System \ref{nn_topopt:mechanical_problem}. The collection of a large dataset from the real-life examples is difficult or even impossible. Therefore, we use synthetic data generated by using Topy \cite{topy} --- an open source solver for 2D and 3D topology optimization, based on SIMP approach.

To generate the dataset we sampled the pseudo-random problem formulations and performed 100 iterations of standard SIMP method. Each problem is defined by the constraints and the loads. The strategy of sampling is the following:

\begin{itemize}
\item The number of nodes with fixed $x$ and $y$ translations and the number of loads is sampled from the Poisson distribution: 
\begin{gather*}
N_{x} \sim P(\lambda = 2),\\
N_{y}, N_{\text{L}} 
\sim P(\lambda =  1)
\end{gather*}
\item The nodes for each of the above-described constraints are sampled from the distribution defined on the grid. The probability of choosing the boundary node is 100 times higher than that for an inner node. 
\item The load values are chosen as $-1$. 
\item The volume fraction is sampled from the Normal distribution 
$f_0 \sim \mathcal{N}(\mu = 0.5, \sigma=0.1)$
\end{itemize}

The obtained dataset\footnote{\label{note_dataset}The dataset and the related code is available at \url{https://github.com/ISosnovik/top}} has 10,000 objects. Each object is a tensor of shape $100 \times 40 \times 40$: 100 iterations of the optimization process for the problem defined on a regular $40 \times 40$ grid. 

\subsection{Training}
We used dataset, described in Section \ref{sect:dataset}, to train our model. During the training process we ``stopped" SIMP solver after $k$ iterations and used the obtained design variables as an input for our model. The input images were augmented with transformations from group D4: horizontal and vertical flips and rotation by 90 degrees. $k$ was sampled from some certain distribution $\mathcal{F}$. Poisson distribution $P(\lambda)$ and discrete uniform distribution $U[1, 100]$ are of interest to us. For training the network we used the objective function of the following form:
\begin{equation}\label{nn_topopt:loss}
\begin{split}
	\mathcal{L} = \mathcal{L}_{\text{conf}} (X_{\text{true}}, X_{\text{pred}}) +  
    \beta \mathcal{L}_{\text{vol}} (X_{\text{true}}, X_{\text{pred}})				
\end{split}
\end{equation}
where the confidence loss is a binary cross-entropy:
\begin{equation}\label{nn_topopt:loss_conf}
	\mathcal{L}_{\text{conf}} (X_{\text{true}}, 	X_{\text{pred}}) = - \frac{1}{NM}
    \sum \limits_{i = 1}^{N} 
    \sum \limits_{j = 1}^{M} \Big[
    X_{\text{true}}^{ij} 
    \log ( X_{\text{pred}}^{ij}) 
    + ( 1 - X_{\text{true}}^{ij})
    \log ( 1 - X_{\text{pred}}^{ij}) \Big]
\end{equation}
where $N \times M$ is the resolution of the image. The second summand in Equation (\ref{nn_topopt:loss}) represents the  volume fraction constraint:
\begin{equation}\label{nn_topopt:loss_const}
	\mathcal{L}_{\text{vol}} (X_{\text{true}}, 	X_{\text{pred}}) =
(\bar{X}_{\text{pred}} - \bar{X}_{\text{true}})^2
\end{equation}

We used ADAM \cite{KingmaADAM:OPTIMIZATION} optimizer with default parameters. We halved the learning rate once during the training process. All code is written in Python\footnote{\label{note_code}The implementation is available at \url{https://github.com/ISosnovik/nn4topopt}}. For neural networks, we used Keras \cite{chollet2015keras} with TensorFlow \cite{tensorflow2015-whitepaper} backend. NVIDIA Tesla K80 was used for deep learning computations. The training of a neural network from scratch took about 80-90 minutes.

\section{Results}
\label{sec:results}

\begin{figure}
    \centering
    \includegraphics[width=0.9\textwidth]
    {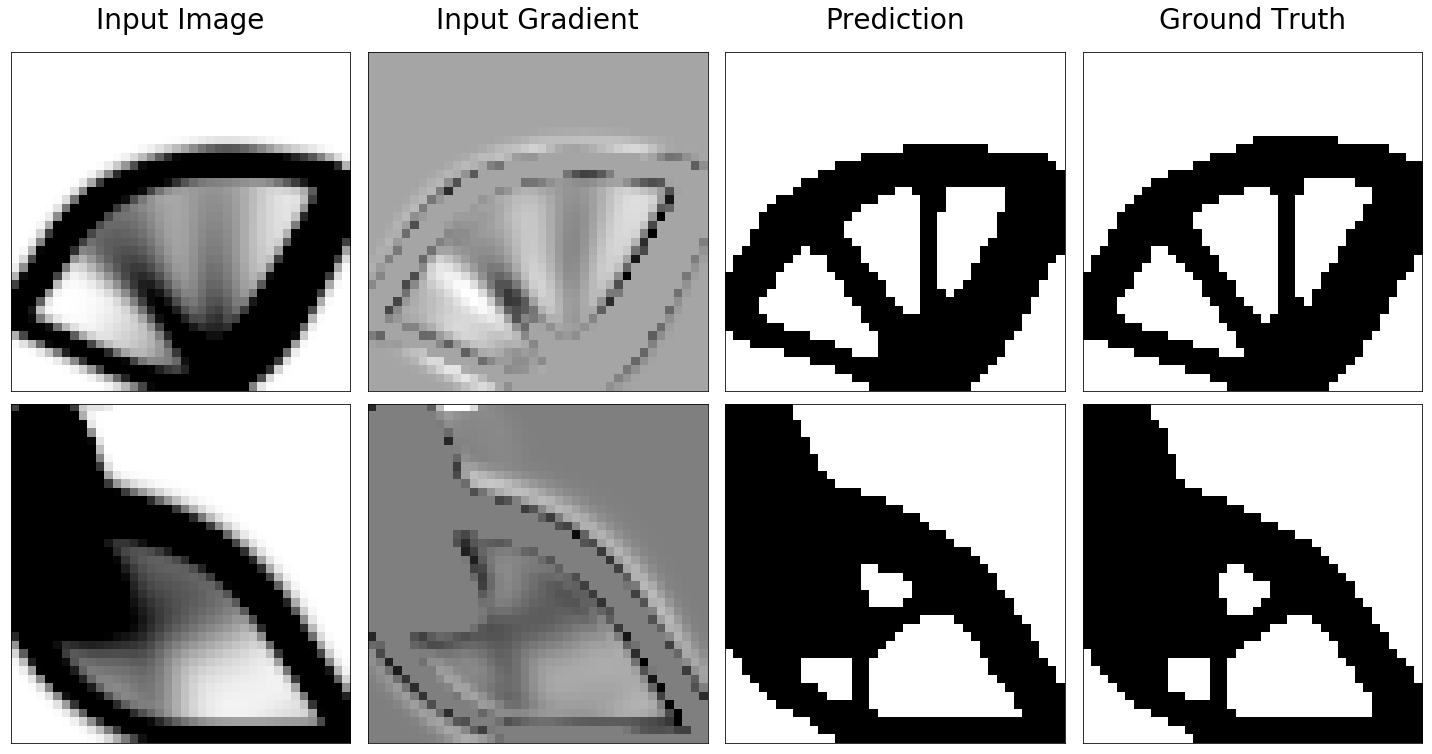}
    \caption[Results of the application of proposed model to the prediction of the final structure.]{Top: SIMP is stopped after 8 iterations, binary accuracy 0.96, mean IoU 0.92; Bottom: solver is stopped after 5 iterations, binary accuracy 0.98, mean IoU 0.95.}
\label{fig:pred_0}
\end{figure}

The goal of our experiments is to demonstrate that the proposed model and the overall pipeline are useful for solving topology optimization problems. We compare the performance of our approach with standard SIMP solver \cite{topy} in terms of the accuracy of the obtained structure and the average time consumption. 

We report two metrics from common image segmentation evaluation: Binary Accuracy and Intersection over Union (IoU). Let $n_l,\;l=0,1$ be the total number of pixels of class $l$. The $\omega_{tp}, \; t, p = 0,1$ is a total number of pixels of class $t$ predicted to belong to class $p$. Therefore:
\begin{equation}\label{nn_topopt:metrics}
	\text{Bin. Acc.} = \frac{\omega_{00} + \omega_{11}}{n_0 + n_1}; \quad
    \text{IoU} = \frac{1}{2}
    \Big[
    \frac{\omega_{00}}{n_0 + \omega_{10}} +
    \frac{\omega_{11}}{n_1 + \omega_{01}}
    \Big]
\end{equation}

We examine 4 neural networks with the same architecture but trained with different policies. The number of iterations after which we ``stopped" SIMP algorithm was sampled from different distributions. We trained one neural network by choosing discrete uniform distribution $U[1, 100]$ and another three models were trained with Poisson distribution $P(\lambda)$ with $\lambda=5, 10, 30$. 

\subsection{Accuracy and performance}

We conducted several experiments to illustrate the results of the application of the proposed pipeline and the exact model to mechanical problems. Figure \ref{fig:pred_0} demonstrates that our neural network restores the final structure while being used even after 5 iterations. The output of the model is close to that of SIMP algorithm. The overall topology of the structure is the same. Furthermore, the time consumption of the proposed method, in this case, is almost 20 times smaller. 

Neural networks trained with different policies produce close results: models preserve the final structure up to some rare pixel-wise changes. However, the accuracy of these models depends on the number of the initial iterations performed by SIMP algorithm. Tables \ref{nn_topopt:table_acc_mech}, \ref{nn_topopt:table_iou_mech} summarize the results obtained in the series of experiments. The trained models demonstrate the sufficiently more accurate results comparing to the thresholding applied after the same number of iterations of SIMP method. Some models benefit when they are applied after 5-10 iterations, while others demonstrate better result in the middle or at the end of the process. The proposed pipeline could significantly accelerate the overall algorithm with minimal reduction in accuracy, especially when CNN is used at the beginning of the optimization process. 

The neural network which used discrete uniform distribution during the training process does not demonstrate the highest binary accuracy and IoU comparing to other models till the latest iterations. However, this model allows one to outperform the SIMP algorithm with thresholding throughout the optimization process.

\begin{table}
\begin{center}
\caption{Binary Accuracy of the proposed method and the standard one on the mechanical dataset.}
\begin{tabular}{ |c|ccccccccc| }
  \hline
  &\multicolumn{9}{ |c| }{Iteration}\\
  \hline
  Method & 5 & 10 & 15 & 20 & 30 & 40 & 50 & 60 & 80 \\
  \hline
  Thresholding & 92.9 & 95.4 & 96.5 & 97.1 & 97.7 & 98.1 & 98.4 & 98.6 & 98.9\\
  CNN $P(5)$ & \textbf{ 95.8} & 97.3 & 97.7 & 97.9 & 98.2 & 98.4 & 98.5 & 98.6 & 98.7\\
  CNN $P(10)$ & 95.4 & \textbf{ 97.6} & \textbf{ 98.1} & 98.4 & 98.7 & 98.9 & 99.0 & 99.0 & 99.0\\
  CNN $P(30)$ & 92.7 & 96.3 & 97.8 & \textbf{ 98.5} & \textbf{ 99.0} & \textbf{ 99.2} & \textbf{ 99.4} & \textbf{ 99.5} & 99.6\\
  CNN $U[1, 100]$ & 94.7 & 96.8 & 97.7 & 98.2 & 98.7 & 99.0 & 99.3 & 99.4 & \textbf{ 99.6}\\
  \hline
\end{tabular}
\label{nn_topopt:table_acc_mech}
\end{center}
\end{table}

\begin{table}
\begin{center}
\caption{Intersection over Union (IoU) of the proposed method and the standard one on the mechanical dataset.}
\begin{tabular}{ |c|ccccccccc| }
  \hline
  &\multicolumn{9}{ |c| }{Iteration}\\
  \hline
  Method & 5 & 10 & 15 & 20 & 30 & 40 & 50 & 60 & 80 \\
  \hline
  Thresholding & 86.8 & 91.2 & 93.3 & 94.3 & 95.6 & 96.3 & 96.8 & 97.3 & 97.9\\
  CNN $P(5)$ & \textbf{ 92.0} & 94.7 & 95.4 & 96.0 & 96.5 & 96.9 & 97.1 & 97.3 & 97.4\\
  CNN $P(10)$ & 91.1 & \textbf{ 95.3} & \textbf{ 96.4} & 96.9 & 97.4 & 97.8 & 98.0 & 98.0 & 98.1\\
  CNN $P(30)$ & 86.4 & 92.9 & 95.7 & \textbf{ 97.0} & \textbf{ 98.1} & \textbf{ 98.5} & \textbf{ 98.8} & \textbf{ 99.0} & 99.2\\
  CNN $U[1, 100]$ & 90.0 & 93.9 & 95.5 & 96.4 & 97.5 & 98.1 & 98.6 & 98.8 & \textbf{ 99.2}\\
  \hline
\end{tabular}
\label{nn_topopt:table_iou_mech}
\end{center}
\end{table}

\subsection{Transferability}
This research is dedicated to the application of neural networks to the topology optimization  of  minimal compliance problems. Nevertheless, the proposed model does not rely on any prior knowledge of the nature of the problem. Despite the fact that we used mechanical dataset during the training, other types of problems from topology optimization framework could be solved by using the proposed pipeline. To examine the generalization properties of our model, we generated the small dataset of heat conduction problems defined on $40 \times 40$ regular grid. The exact solution and the intermediate densities for the problems were obtained in absolutely the same way as it was described in Section \ref{sec:learn_topopt}. 

The conducted experiments are summarized in Table \ref{nn_topopt:table_acc_heat}, \ref{nn_topopt:table_iou_heat}. During the initial part of the optimization process, the results of the pre-trained CNNs are more accurate than this of thresholding. Our model approximates the mapping to the final structure precisely when the training dataset and the validation dataset are from the same distribution. However, it mimics updates of SIMP method during the initial iterations even when CNN is applied to another dataset. Therefore, this pipeline could be useful for the fast prediction of the rough structure in various topology optimization problems.

\begin{figure}
    \includegraphics[width=1.0\textwidth]
    {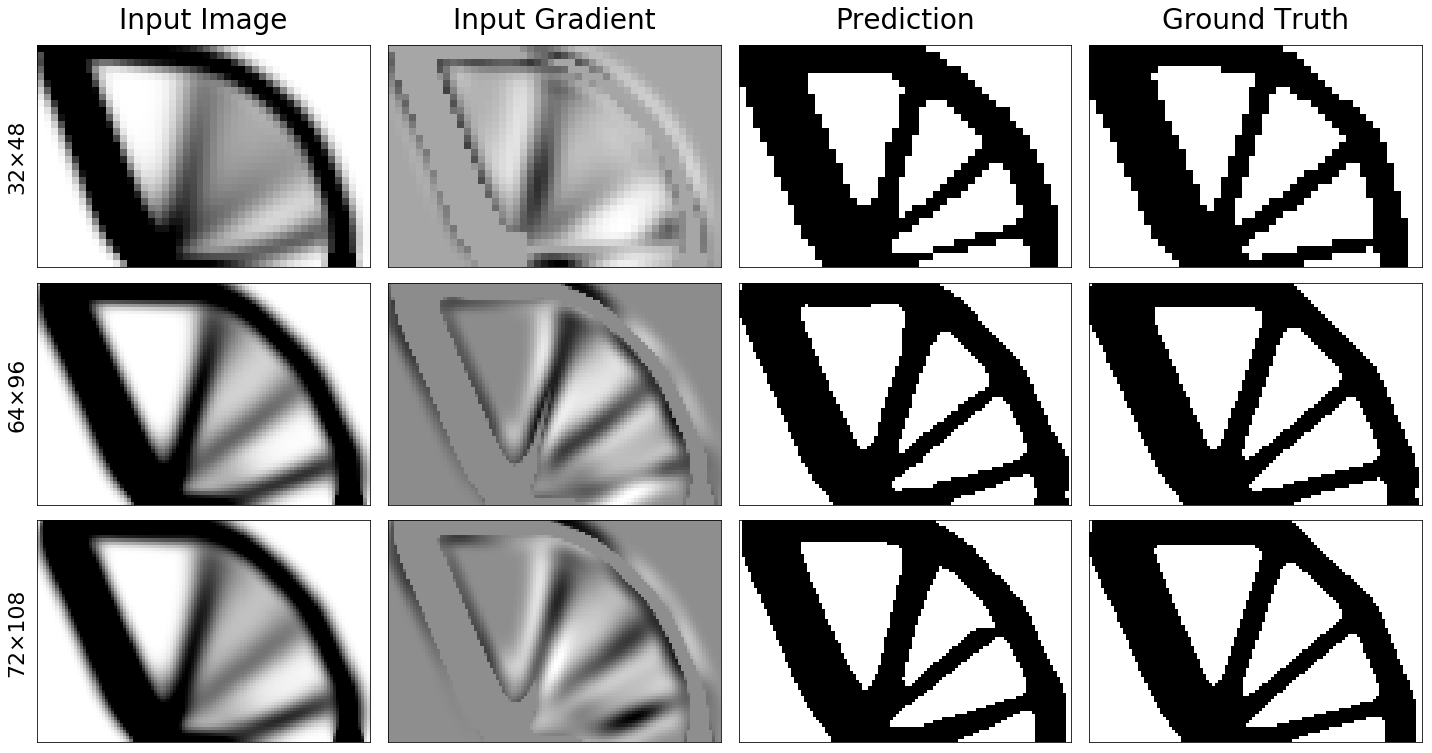}
    \caption[]{Results of the application of the proposed CNN to the problems defined on grids with resolution and aspect ratio different from that of the training dataset.}
\label{fig:res_resolution}
\end{figure}

The neural network described in Section \ref{sec:learn_topopt} is fully-convolutional, i.e. it consists of Convolutional, Pooling, Upsampling and Dropout layers. The architecture itself does not have any constraints on the size of the input data. In this experiment, we checked the scalable properties of our method. The model we examined had been trained on the original dataset with square images of size $40 \times 40$. Figure \ref{fig:res_resolution} visualizes the result of the application of CNN to the problems defined on grids with different resolution. Here we can see that changes in the aspect ratio and reasonable changes in the resolution of the input data do not affect the accuracy of the model. Pre-trained neural network successfully reconstructs the final structure for a given problem. Significant changes of the size of the input data require additional training of the model because the typical size of a common patterns changes with the increase of the resolution of an image. Nevertheless, demonstrated cases did not require one to tune neural network and allowed to transfer model from one resolution to another.

\begin{table}
\begin{center}
\caption{Binary Accuracy of the proposed method and the standard one on heat conduction dataset. Models were trained on the minimal compliance dataset.}
\begin{tabular}{ |c|ccccccccc| }
  \hline
  &\multicolumn{9}{ |c| }{Iteration}\\
  \hline
  Method & 5 & 10 & 15 & 20 & 30 & 40 & 50 & 60 & 80 \\
  \hline
  Thresholding & 97.5 & 98.4 & 98.8 & 99.1 & 99.4 & \textbf{ 99.6} & \textbf{ 99.7} & \textbf{ 99.8} & \textbf{ 99.9}\\
  CNN $P(5)$ & 98.1 & 98.7 & 99.0 & 99.2 & 99.4 & 99.5 & 99.6 & 99.7 & 99.7\\
  CNN $P(10)$ & \textbf{ 98.1} & 98.8 & 99.0 & 99.2 & 99.4 & 99.5 & 99.6 & 99.7 & 99.8\\
  CNN $P(30)$ & 97.3 & \textbf{ 99.0} & \textbf{ 99.2} & \textbf{ 99.4} & \textbf{ 99.5} & 99.6 & 99.7 & 99.7 & 99.8\\
  CNN $U[1, 100]$ & 97.8 & 98.8 & 99.1 & 99.3 & 99.5 & 99.6 & 99.7 & 99.7 & 99.8\\
  \hline
\end{tabular}
\label{nn_topopt:table_acc_heat}
\end{center}
\end{table}

\begin{table}
\begin{center}
\caption{IoU of the proposed method and the standard one on heat conduction dataset. Models were trained on the minimal compliance dataset.}
\begin{tabular}{ |c|ccccccccc| }
  \hline
  &\multicolumn{9}{ |c| }{Iteration}\\
  \hline
  Method & 5 & 10 & 15 & 20 & 30 & 40 & 50 & 60 & 80 \\
  \hline
  Thresholding & 95.1 & 96.8 & 97.6 & 98.1 & 98.8 & \textbf{ 99.2} & \textbf{ 99.4} & \textbf{ 99.6} & \textbf{ 99.9}\\
  CNN $P(5)$ & 96.2 & 97.5 & 98.0 & 98.4 & 98.8 & 99.0 & 99.2 & 99.3 & 99.5\\
  CNN $P(10)$ & \textbf{ 96.3} & 97.6 & 98.1 & 98.4 & 98.9 & 99.1 & 99.3 & 99.4 & 99.5\\
  CNN $P(30)$ & 94.8 & \textbf{ 98.0} & \textbf{ 98.5} & \textbf{ 98.7} & \textbf{ 99.0} & 99.2 & 99.3 & 99.4 & 99.5\\
  CNN $U[1, 100]$ & 95.7 & 97.7 & 98.2 & 98.6 & 98.9 & 99.2 & 99.3 & 99.4 & 99.6\\
  \hline
\end{tabular}
\label{nn_topopt:table_iou_heat}
\end{center}
\end{table}

\section{Related work}
\label{sec:related}

The current research is supposed to be the first one which utilizes deep learning approach for the topology optimization problem. It is inspired by the recent successful application of deep learning to problems in computational physics. Greff at al. \cite{Greff2016} used the fully-connected neural  network as a  mapping function from the nano-material configuration and the input voltage to the output current.  The adaptation of restricted Boltzmann machine  for solving the Quantum Many-Body Problem was demonstrated  in paper \cite{CarleoSolvingNetworks}. Mills et al.  \cite{MillsK.SpannerM.DeepEquation} used the machinery of deep learning to learn the mapping between potential and energy, bypassing the need to numerically solve the Schr\"{o}dinger equation and the need for computing wave functions.  Tompson et al. \cite{Tompson2016AcceleratingNetworks} and  Kiwon et al. \cite{Um2017LiquidNetworks} accelerated the process of modeling of liquids  by the application of neural networks. The paper \cite{SmithANI-1:Cost} demonstrates how a deep neural network trained on quantum mechanical density functional theory calculations can learn an accurate and transferable potential for organic molecules. The cutting-edge research \cite{Paganini2017CaloGAN:Networks} shows how generative adversarial networks could be used  for simulating 3D high-energy particle showers in multi-layer electromagnetic calorimeters. 
\section{Conclusion}
\label{sec:conclusion}

In this paper, we proposed a neural network as an effective tool for the acceleration of topology optimization process. Out model learned the mapping from the intermediate result of the iterative method to the final structure of the design domain. It allowed us to stop SIMP method earlier and significantly decrease the total time consumption.

We demonstrated that the model trained on the dataset of minimal compliance problems could produce the rough approximation of the solution for other types of topology optimization problems. Various experiments showed that the proposed neural network transfers successfully from the dataset with a small resolution to the problems defined on the grids with better resolution.

\medskip
\newpage
\bibliographystyle{elsarticle-num}
\bibliography{biblio}

\begin{thebibliography}{10}
\expandafter\ifx\csname url\endcsname\relax
  \def\url#1{\texttt{#1}}\fi
\expandafter\ifx\csname urlprefix\endcsname\relax\def\urlprefix{URL }\fi
\expandafter\ifx\csname href\endcsname\relax
  \def\href#1#2{#2} \def\path#1{#1}\fi

\bibitem{topopt_application}
M.~P. Bends{\o}e, E.~Lund, N.~Olhoff, O.~Sigmund, Topology
  optimization-broadening the areas of application, Control and Cybernetics 34
  (2005) 7--35.

\bibitem{Bendsoe1989}
M.~P. Bends{\o}e, Optimal shape design as a material distribution problem,
  Structural and multidisciplinary optimization 1~(4) (1989) 193--202.

\bibitem{beso_1}
C.~Mattheck, S.~Burkhardt, A new method of structural shape optimization based
  on biological growth, International Journal of Fatigue 12~(3) (1990)
  185--190.

\bibitem{beso_2}
Y.~M. Xie, G.~P. Steven, A simple evolutionary procedure for structural
  optimization, Computers \& structures 49~(5) (1993) 885--896.

\bibitem{Mlejnek1992}
H.~Mlejnek, Some aspects of the genesis of structures, Structural and
  Multidisciplinary Optimization 5~(1) (1992) 64--69.

\bibitem{bendsoe1995optimization}
M.~P. Bends{\o}e, Optimization of structural topology, shape, and material,
  Vol. 414, Springer, 1995.

\bibitem{sigmund1997}
O.~Sigmund, On the design of compliant mechanisms using topology optimization,
  Journal of Structural Mechanics 25~(4) (1997) 493--524.

\bibitem{bourdin2001}
B.~Bourdin, Filters in topology optimization, International Journal for
  Numerical Methods in Engineering 50~(9) (2001) 2143--2158.

\bibitem{Svanberg2013}
K.~Svanberg, H.~Sv{\"a}rd, Density filters for topology optimization based on
  the {Pythagorean} means, Structural and Multidisciplinary Optimization 48~(5)
  (2013) 859--875.

\bibitem{Groenwold2009}
A.~A. Groenwold, L.~Etman, A simple heuristic for gray-scale suppression in
  optimality criterion-based topology optimization, Structural and
  Multidisciplinary Optimization 39~(2) (2009) 217--225.

\bibitem{topopt99lines}
O.~Sigmund, A 99 line topology optimization code written in {Matlab},
  Structural and multidisciplinary optimization 21~(2) (2001) 120--127.

\bibitem{topopt88lines}
E.~Andreassen, A.~Clausen, M.~Schevenels, B.~S. Lazarov, O.~Sigmund, Efficient
  topology optimization in {MATLAB} using 88 lines of code, Structural and
  Multidisciplinary Optimization 43~(1) (2011) 1--16.

\bibitem{topy}
W.~Hunter, et~al., Topy - topology optimization with python,
  \url{https://github.com/williamhunter/topy} (2017).

\bibitem{RonnebergerU-Net:Segmentation}
O.~Ronneberger, P.~Fischer, T.~Brox, U-net: Convolutional networks for
  biomedical image segmentation, in: International Conference on Medical Image
  Computing and Computer-Assisted Intervention, Springer, 2015, pp. 234--241.

\bibitem{dropout}
G.~E. Hinton, N.~Srivastava, A.~Krizhevsky, I.~Sutskever, R.~R. Salakhutdinov,
  Improving neural networks by preventing co-adaptation of feature detectors,
  arXiv preprint arXiv:1207.0580.

\bibitem{KingmaADAM:OPTIMIZATION}
D.~Kingma, J.~Ba, Adam: A method for stochastic optimization, arXiv preprint
  arXiv:1412.6980.

\bibitem{chollet2015keras}
F.~Chollet, et~al., Keras, \url{https://github.com/fchollet/keras} (2015).

\bibitem{tensorflow2015-whitepaper}
M.~Abadi, A.~Agarwal, P.~Barham, E.~Brevdo, Z.~Chen, C.~Citro, G.~S. Corrado,
  A.~Davis, J.~Dean, M.~Devin, et~al., Tensorflow: Large-scale machine learning
  on heterogeneous distributed systems, arXiv preprint arXiv:1603.04467.

\bibitem{Greff2016}
K.~Greff, R.~van Damme, J.~Koutnik, H.~Broersma, J.~Mikhal, C.~Lawrence,
  W.~van~der Wiel, J.~Schmidhuber, Using neural networks to predict the
  functionality of reconfigurable nano-material networks.

\bibitem{CarleoSolvingNetworks}
G.~Carleo, M.~Troyer, Solving the quantum many-body problem with artificial
  neural networks, Science 355~(6325) (2017) 602--606.

\bibitem{MillsK.SpannerM.DeepEquation}
K.~Mills, M.~Spanner, I.~Tamblyn, Deep learning and the {Schr\"{o}dinger}
  equation, arXiv preprint arXiv:1702.01361.

\bibitem{Tompson2016AcceleratingNetworks}
J.~Tompson, K.~Schlachter, P.~Sprechmann, K.~Perlin, Accelerating {Eulerian}
  fluid simulation with convolutional networks, arXiv preprint
  arXiv:1607.03597.

\bibitem{Um2017LiquidNetworks}
K.~Um, X.~Hu, N.~Thuerey, Liquid splash modeling with neural networks, arXiv
  preprint arXiv:1704.04456.

\bibitem{SmithANI-1:Cost}
J.~S. Smith, O.~Isayev, A.~E. Roitberg, Ani-1: an extensible neural network
  potential with {DFT} accuracy at force field computational cost, Chemical
  Science 8~(4) (2017) 3192--3203.

\bibitem{Paganini2017CaloGAN:Networks}
M.~Paganini, L.~de~Oliveira, B.~Nachman, Calogan: Simulating {3D} high energy
  particle showers in multi-layer electromagnetic calorimeters with generative
  adversarial networks, arXiv preprint arXiv:1705.02355.

\end{thebibliography}

\cleardoublepage
\appendix
\section{Dataset}

\begin{figure}[ht]
    \centering
   \includegraphics[width=0.98\textwidth]
    {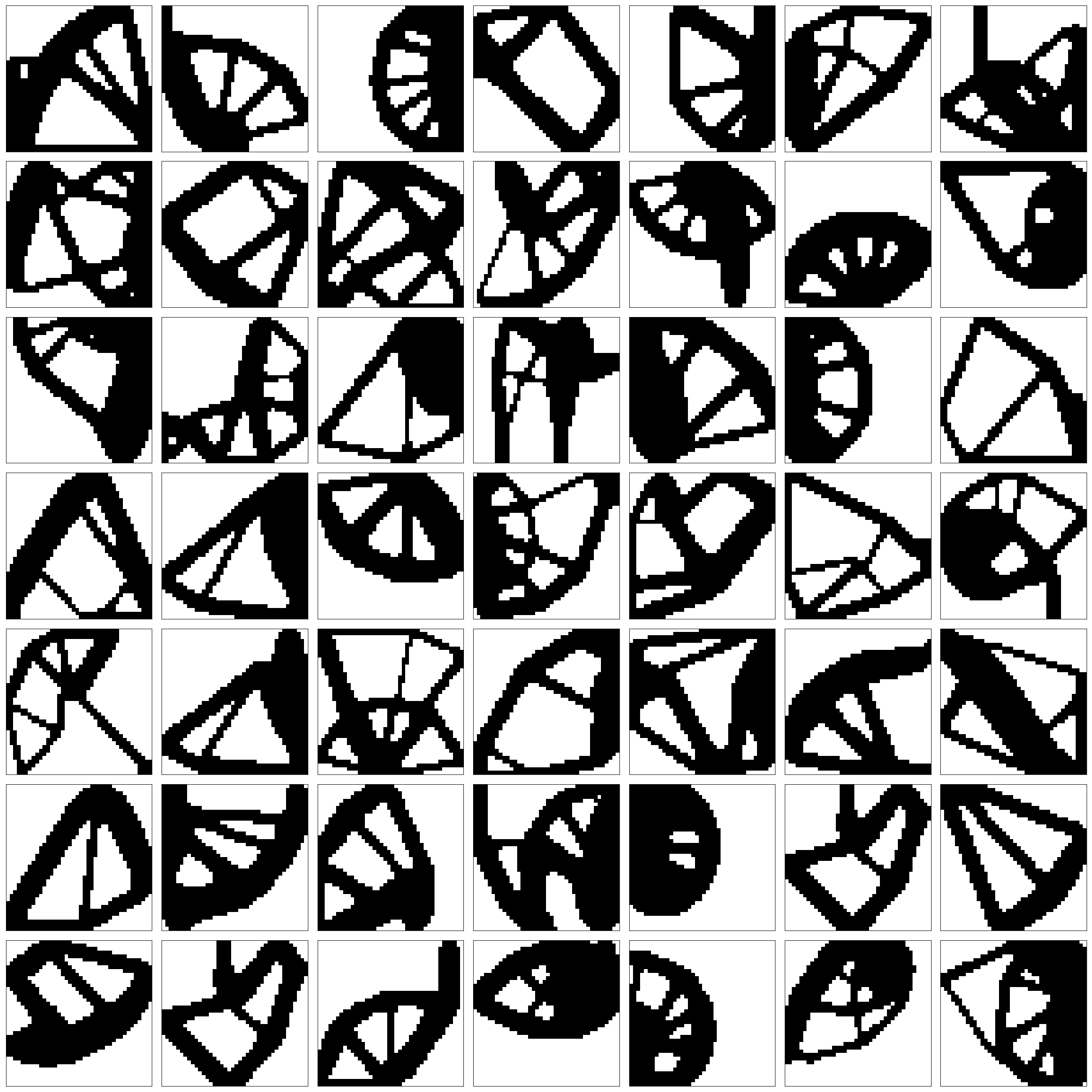}
    \caption[The dataset of topology optimization process]{The samples from the dataset of topology optimization process.}
\label{fig:dataset}
\end{figure}

\newpage
\section{Results}

\begin{figure}[ht]
    \centering
   \includegraphics[width=0.9
\textwidth]
    {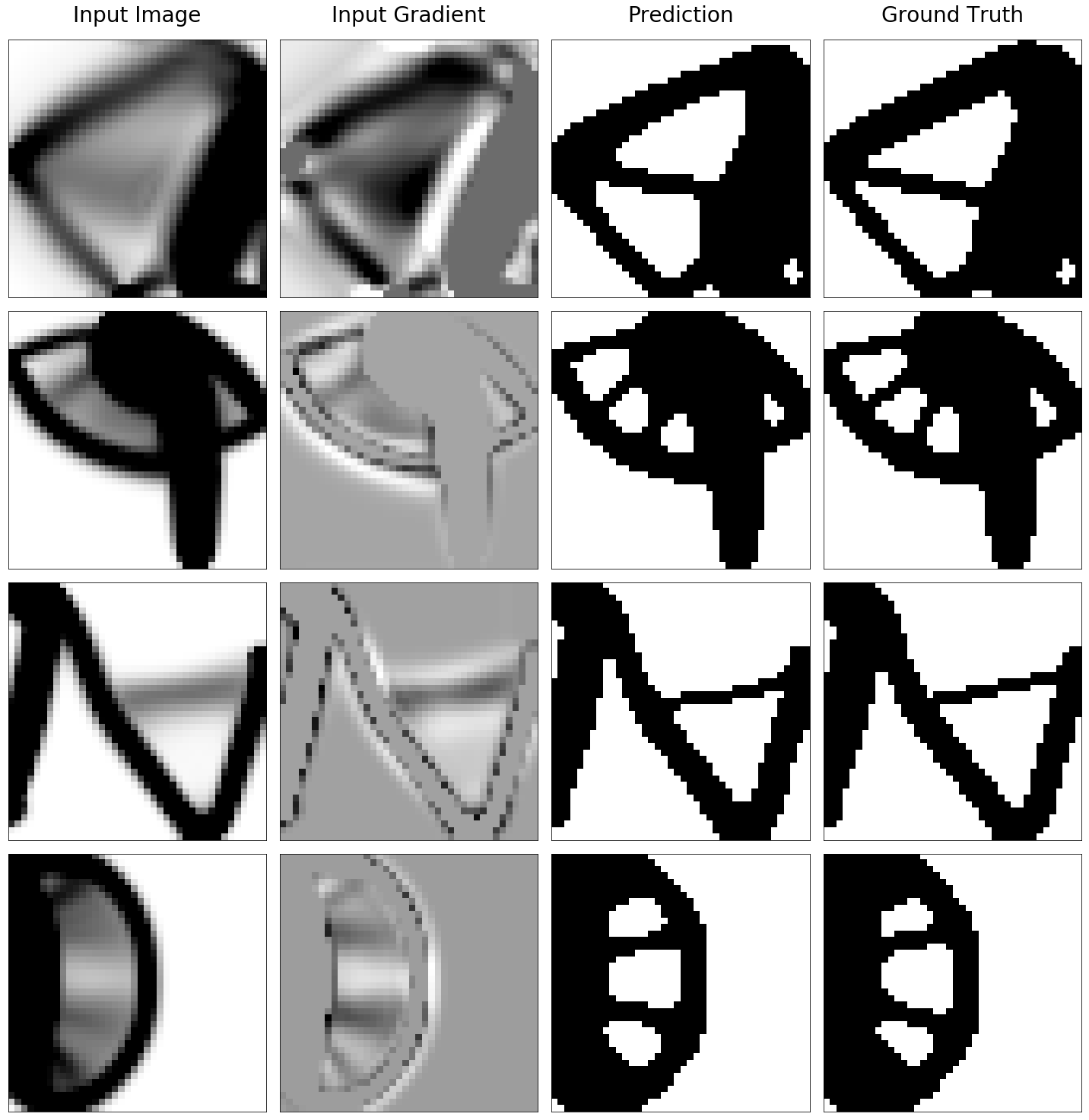}
    \caption{Results of the application of the proposed model to the prediction of the final structure.}
\label{fig:pred_appendix_big}
\end{figure}

\end{document}